\UseRawInputEncoding
\documentclass[lettersize,journal]{IEEEtran}
\usepackage{amsmath,amsfonts}
\usepackage{algorithmic}
\usepackage{algorithm}
\usepackage{array}
\usepackage[caption=false,font=normalsize,labelfont=sf,textfont=sf]{subfig}
\usepackage{textcomp}
\usepackage{stfloats}
\usepackage{url}
\usepackage{verbatim}
\usepackage{graphicx}
\usepackage{esvect}
\usepackage{cancel}
\usepackage[noadjust]{cite}

\usepackage{lipsum}
\usepackage{graphicx}
\usepackage{siunitx}
\newcommand{\marker}[1]{\text{M}_{\text{#1}}}

\begin{document}

\title{Quantifying Ergonomics in the \textit{Elevate}  Soft Robotic Suit}

\author{Peter Bryan, \textit{Student Member, IEEE,} Rejin John Varghese, \textit{Member, IEEE,} and Dario Farina, \textit{Fellow, IEEE}
\thanks{This research is supported in part by the UK EPSRC EP/T020970/1 NISNEM grant.} 
\thanks{All authors are with the Department of Bioengineering, Imperial College of Science, Technology and Medicine, London W12 0BZ, UK (email: 
 {\tt\footnotesize \{peter.bryan21,r.varghese15,,d.farina\}@imperial.ac.uk}
}
}




\maketitle

\begin{abstract}
Soft robotic suits have the potential to rehabilitate, assist, and augment the human body. The low weight, cost, and minimal form-factor of these devices make them ideal for daily use by both healthy and impaired individuals. However, challenges associated with data-driven, user-specific, and comfort-first design of human-robot interfaces using soft materials limit their widespread translation and adoption.
In this work, we present the quantitative evaluation of ergonomics and comfort of the \textit{Elevate} suit - a cable driven soft robotic suit that assists shoulder elevation. Using a motion-capture system and force sensors, we measured the suit's ergonomics during assisted shoulder elevation up to $\SI{70}{\degree}$.  Two 4-hour sessions were conducted with one subject, involving transmitting cable tensions of up to $\SI{200}{\newton}$ with no discomfort reported. We estimated that the pressure applied to the shoulder during assisted movements was within the range seen in a human grasp ($\approx\SI{69.1-85.1}{\kilo\pascal}$), and approximated a volumetric compression of $<3\%$ and $<8\%$ across the torso and upper arm respectively. These results validate \textit{Elevate}'s ergonomic design in preparation for future studies with patient groups. 

\end{abstract}

\begin{IEEEkeywords}
Wearable Robotics, Soft Robot Materials and Design, Prosthetics and Exoskeletons, Physically Assistive Devices, Physical Human-Robot Interaction.
\end{IEEEkeywords}

\section{Introduction}
Disorders such as stroke, spinal cord injury, muscular dystrophy, and peripheral nerve injury, that affect movement can have a serious negative impact on the quality of life of those affected \cite{bos_prevalence_2019}. For individuals with upper-limb impairment, daily tasks such as eating, writing, and handling objects may become challenging or even impossible. Devices that can support the arm, either for ongoing daily assistance or to aid rehabilitation exercises can help alleviate the burden of these impairments \cite{lee_effectiveness_2022}. Similar devices could also be used to augment healthy individuals by reducing loading on the body and increasing endurance during repeated manual tasks. \cite{piao_development_2023}

Soft Robotic Suits (SRSs) are clothing-like robotic devices that wrap around a person's body and work in parallel with their muscles \,\cite{xiloyannis_soft_2022}. The SRS field has grown significantly over the past decade, with notable advances in control strategies, actuation technologies, and suit architecture \cite{xiloyannis_soft_2022}. Despite this progress, key challenges have prevented the adoption of SRSs for rehabilitation or as assistive devices \cite{varghese_wearable_2018}.

SRSs also known as exosuits and soft exoskeletons are typically constructed from flexible materials such as foams, fabrics, inextensible webbing, and plastics. This gives SRSs their characteristic benefits (lightweight, compliant, low form, and non-restrictive nature) but also raises design challenges that prevent them from achieving the same levels of force assistance as their rigid counterparts. Design optimisation to minimise mechanical losses and reduce forces on the wearer while maximising assistive capabilities is crucial for widespread real-world adoption. This becomes especially important in applications involving wearers with muscular disorders or weakness, as they are often more susceptible to adverse effects from improperly applied external forces \cite{tiffreau_pain_2006,qu_observation_2025}. 
By integrating intuitive, iterative design with data-driven decision-making and quantitative analysis, the development of an SRS that advances current assistive capabilities and enables real-world use becomes feasible.

A limited number of studies have attempted to quantify the physical interaction between SRSs and the wearer. Piao et al. \cite{piao_development_2023} evaluated suit-body interaction through surveys and garment pressure measurements at multiple body locations, relating measured interface pressures to subjective comfort during assisted gait. Similarly, Xiloyannis et al. \cite{xiloyannis_characterisation_2018} used a high density pressure mat to measure spatially resolved contact pressure with the aim to inform interface design and improve wearer comfort of an elbow SRS. While extensive pressure measurements such as those made with high-density capacitative pressure sensing mats would be ideal for understanding suit-body interactions, their cost makes them prohibitive, highlighting the need for new approaches and indirect estimation of these forces using more accessible force and kinematic sensing methods. Yandell et al. \cite{yandell_physical_2017} used a motion-capture system and load cells to quantify suit-body interaction in order to study power transmission and efficiency, demonstrating that suit-body dynamics can significantly affect device effectiveness. These studies represent the primary efforts to go beyond outcome-based assessments and quantify suit-body interaction in soft robotic suits in more detail. 


While this area of research is growing, there remains a lack of quantitative evaluation under dynamic loading, making objective comparison between SRS designs difficult and hindering progress toward real-world deployment. In this work, we present a quantitative evaluation of suit-body interaction forces experienced during use of \textit{Elevate}, a cable-driven soft robotic suit designed to assist shoulder elevation, using motion capture and force sensing to assess ergonomics, comfort, and long-term wearability.

\begin{figure*}
    \centering
    \includegraphics[width=0.95\linewidth]{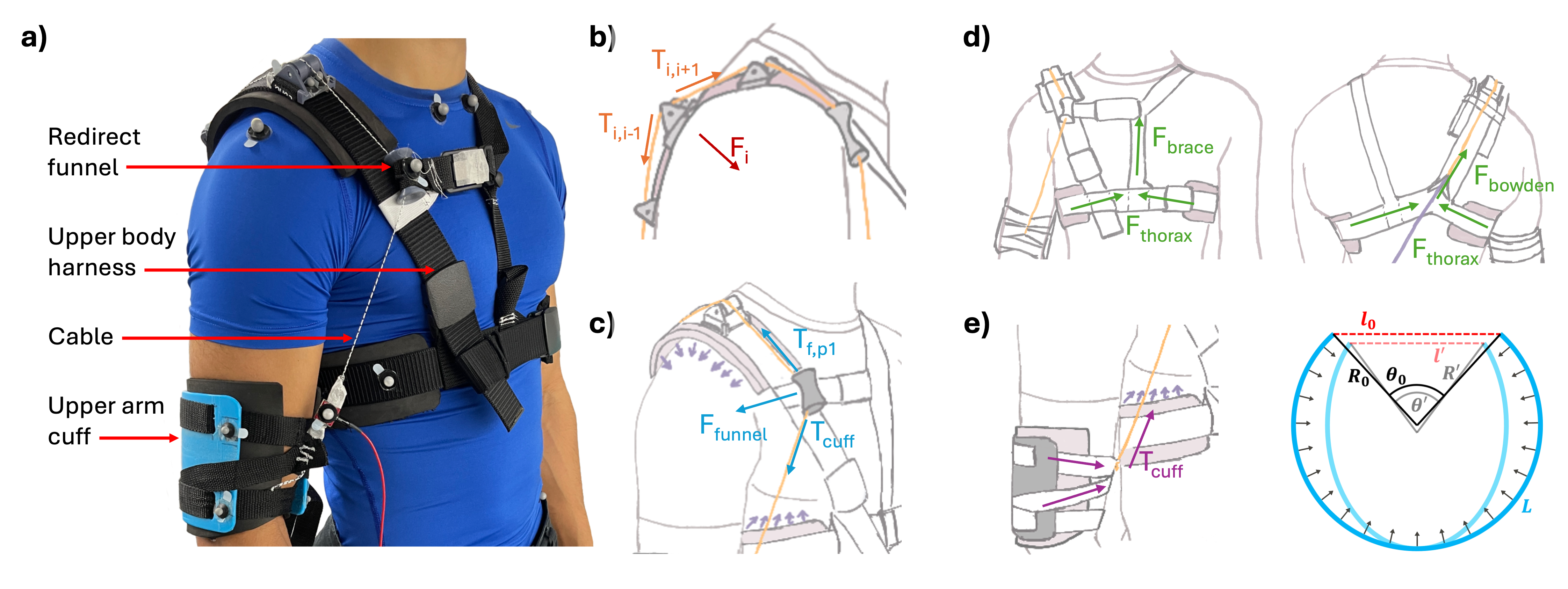}
        \caption{\textbf{a)} The \textit{Elevate} suit. \textbf{b-e)} Estimation of suit-body interaction forces across multiple regions, derived from kinematic and geometric measurements obtained from reflective motion-capture markers, together with cable tension data from force sensors.}
    \label{fig:methods}
\end{figure*}

\section{The Elevate Suit}

The \textit{Elevate} SRS (Fig.\ref{fig:methods}a) consists of a lightweight upper-body harness worn around the torso and shoulders. Assistance forces are transmitted  via a Bowden cable from an off-body Brushless Direct Current (BLDC) motor, over the assisted shoulder to a cuff on the upper arm. The suit is adjustable to accommodate a wide range of body sizes. For this study, actuation and control were implemented using a custom Python-based control interface.

\section{Evaluation Methods}


The methods introduced here use load cells and motion-capture data from reflective markers placed on both the suit and body, at anatomical landmarks as well as other regions of interest (Fig.\,\ref{fig:methods} and Fig.\,\ref{fig:study}, details in section \ref{sec:setup}). To evaluate shoulder elevation (coupled flexion and adduction) induced by \textit{Elevate}, shoulder kinematics were computed by first identifying the plane of elevation and then calculating the elevation angle in that plane at each time instant during the assisted movement. 

At its core, SRS design involves balancing comfort and safety with assistance capability. In the \textit{Elevate} suit, as the motor pulls on the cable, the resulting force is grounded through different parts of the harness (constricting the torso), transmitted over the pulleys (pushing down on the top of the shoulder), and finally applied at the cuff (compressing the arm). We sought to quantify the suit-body interaction at each of these regions with the following metrics:

\begin{enumerate}
    \item normal force and pressure on the shoulder (Fig.\,\ref{fig:methods}b\&c), 
    \item compression across the torso (Fig.\,\ref{fig:methods}d), and
    \item compression across the upper arm (Fig.\,\ref{fig:methods}e).
\end{enumerate}

\subsection{Pressure and Force on the Shoulder}

To estimate normal force and pressure on the shoulder, markers placed along the cable transmission path (Fig.\,\ref{fig:methods}b, Fig.\,\ref{fig:study}a) were used to reconstruct cable segment geometry. Markers at the Bowden cable exit ($\marker{B}$) and distal force sensor ($\marker{FSC}$) defined the endpoints of the exposed cable path. Markers on the pulleys allowed us to compute virtual contact points of the cable on the pulleys  ($\marker{P1-3}$). These together with the redirect funnel ($\marker{F}$) defined the straight cable sections. The net normal force ($\vec{\mathbf{F}}_i$) and pressure ($P$) acting on the body under pulley $i$ is then  calculated to be:
\begin{equation}
    \begin{gathered}
            \vec{\mathbf{F}}_i = \vec{\mathbf{T}}_{i,i+1} + \vec{\mathbf{T}}_{i,i-1} \\
            P_i = \frac{|\vec{\mathbf{F}}_i|}{A_{\text{base}}}
        \end{gathered}
\end{equation}
, where $\vec{\mathbf{T}}_{i,j} = T\hat{\mathbf{n}}_{ij}$ where $T$ is the tension in the cable and $\hat{\mathbf{n}}_{ij}$ represents the unit vector of the cable section, measured originating from pulley $i$ towards pulley $j$.

Fig.\,\ref{fig:body interaction}a presents the estimated pressure experienced by the wearer at the base of each pulley and the force experienced at the funnel (Fig.\,\ref{fig:methods}c). The above computations assume that the friction in the pulleys and funnel were negligible, resulting in approximately equal tension in all segments of the exposed cable. The deformation of the strap and padding foam are also ignored, thus the true pressure would be lower than estimated due to the larger contact area. 

\subsection{Compression across the Torso}

Next, to estimate the compression of the torso, the change in torso volume at the chest harness is approximated (Fig.\,\ref{fig:methods}d). This volume is proportional to the cross sectional area (modelling the torso at the harness to be an elliptical cylinder). This area was approximated as the surface area of the quadrilateral described by four markers around the torso ($\marker{TU1-4}$). This area at every frame is normalised to the area when the suit is unpowered.
Fig.\,\ref{fig:body interaction}b shows the percentage constriction of the torso caused by forces through the chest strap.  In the future, we aim to improve this estimation by increasing the number of markers used.

\subsection{Compression across the Upper Arm}


To quantify upper-arm constriction caused by cuff movement under cable tension, as was done for the torso, the percentage change in area (proportional to cuff volume) is estimated (Fig.\,\ref{fig:methods}e). This area is modelled as a partially open cylinder with constant arc length $L$, radius $R$, and open chord length $l = |\vec{\marker{C1}} - \vec{\marker{C2}}|$. Geometrically, at any instant, this gives the following:

\begin{equation}
    \begin{aligned}
        \frac{l}{2R} = sin(\theta) \text{ and } 
        \frac{L}{2\pi R} = \frac{2\pi - 2\theta}{2\pi}
    \end{aligned}
\end{equation}
, which reduces to the non-linear equation
\begin{equation}
    \frac{l}{2R} = sin(\frac{L}{2R})
\end{equation}
which is solved numerically to estimate the cuff radii ($R_0$, $R'$) in the unpowered and assisted instances. The resulting fractional reduction in volume due to constriction then becomes:

\begin{equation}
    \Delta V = 1- (\frac{R'}{R_0})^2
\end{equation}

\begin{figure*}
    \centering
    \includegraphics[width=\linewidth]{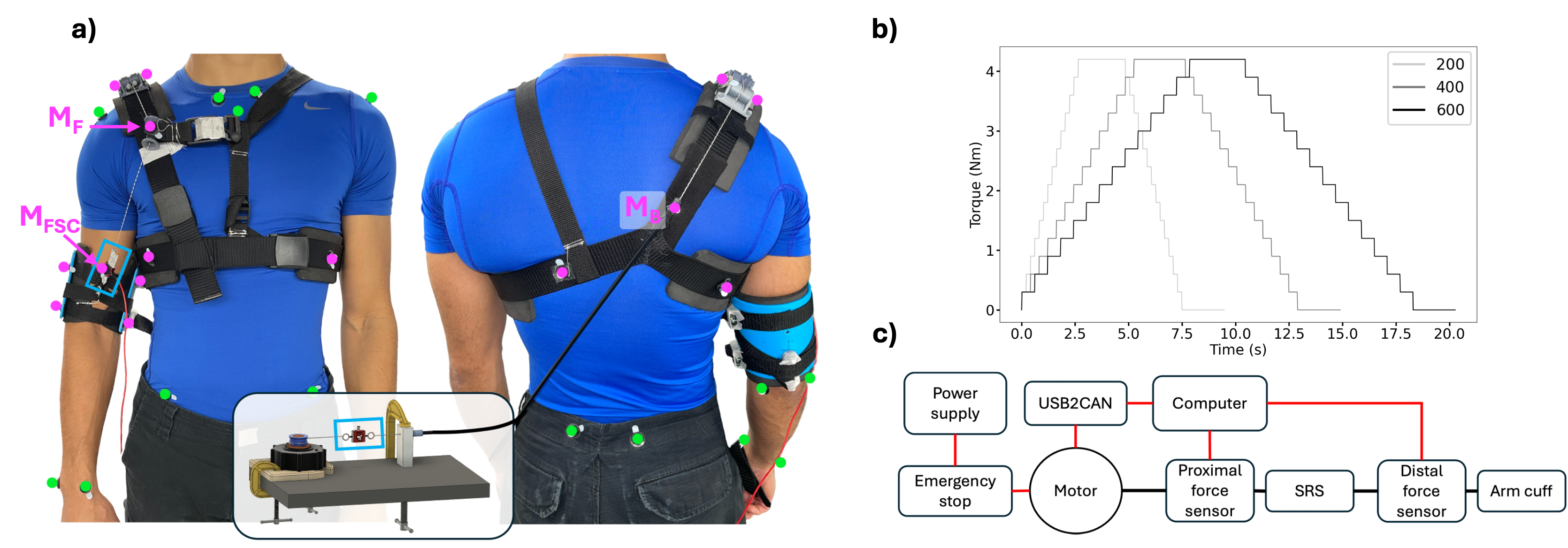}
    \caption{Study setup and protocol. \textbf{a)} Actuation and sensing: locations of the 13 body markers (green) and 17 suit markers (pink). Also shown are the two force sensors (outlined in blue) and motor. \textbf{b)} Motor torque command profile used (three of six trials shown). \textbf{c)} Schematic of the control, actuation, and sensing components used in the study. }
    \label{fig:study}
\end{figure*}
\section{Experimental Setup and Protocol}
\subsection{Experimental Setup}
\label{sec:setup}
A motion-capture system and two force sensors were used to record body kinematics, suit geometry and movement, and cable tension at two points during a series of upper-limb movements performed under the action of the \textit{Elevate} SRS.

For this study, the driving motor was mounted to a table, with the Bowden cable housing fixed to the surface \SI{100}{\milli\meter} from the motor. This setup created a section of exposed cable where the inline LSB205 load cell (Futek, CA, USA) was installed (Fig.\,\ref{fig:study}a\&c). The motor was powered by a $\SI{24}{\volt}$, $\SI{10}{\ampere}$ DC power supply equipped with an emergency cut-off button accessible to the wearer. The signal from each force sensor was amplified using the IAA100 load-cell amplifiers (Futek, CA, USA) before being sampled via an Arduino Uno (Arduino, Italy) and transmitted to the laptop. The proximal force sensor, which is mobile and translates with the cable, was placed between the motor and the start of the Bowden cable housing to measure the input cable tension while the distal force sensor was attached at the cuff to capture the force applied on the arm (Fig.\,\ref{fig:study}a\&c).

The study was performed on a single healthy participant (male, age: 26, height: $\SI{176}{\centi\meter}$, weight: $\SI{75}{\kilo\gram}$) wearing the \textit{Elevate} SRS. All experimental procedures adhered to the ethical guidelines set by the Imperial College Research Ethics Committee (ICREC) under approved application~22IC7655. For the following shoulder elevation tasks, the motor was operated in current control mode to ensure the wearer's safety. Prior to the study, a maximum motor torque value of \SI{4.2}{\newton\meter} was identified empirically such that the torque would cause full cable travel without injury to the wearer, this resulted in approximately $\SI{200}{\newton}$ cable tension at the proximal sensor. 
Retro-reflective markers were placed on 13 anatomical landmarks on the body of the participant with an additional 17 markers placed at key locations on the suit (Fig.\,\ref{fig:study}a) based on the evaluation methods previously described. The motion-capture system used was the Vicon Vero (Vicon, UK) with data captured at $\SI{100}{\hertz}$. All post-processing including reconstruction, gap filling, and filtering was done in the Vicon Nexus software. The trajectory data was transformed into a local coordinate system established using the markers placed at the clavicle of the subject.

\subsection{Experimental Protocol}
\label{sec:protocol}
As shoulder elevation requires work against gravity, steps up or down in motor torque caused the arm to rise or fall by a small angle until equilibrium was reached. The resulting shoulder angle was a function of cable tension, suit measurements, arm weight, and internal losses. This approach enabled controlled progression through shoulder elevation angle while ensuring the motor torque stayed within predetermined safety limits. To evaluate suit-body interaction, the following shoulder elevation task was performed. 

The participant was instructed to keep their arm fully relaxed while the motor followed a predefined trapezoidal torque profile. After initial pre-tensioning, the motor torque command progressively increased from $\SI{0.0}{\newton\meter}$ to $\SI{4.2}{\newton\meter}$ and then decreased back to $\SI{0.0}{\newton\meter}$ in steps of $\SI{0.3}{\newton\meter}$ (Fig.\,\ref{fig:study}b). Three repetitions of this profile constituted a single trial. During each trial, motion-capture data, motor parameters, and two force sensor signals were recorded. In total, six trials were performed, each with a different rate of torque change allowing variability due to varying dynamics to also be assessed. This resulted in the arm being lifted from neutral position by the side of the body to the maximum elevation angle ($\approx\SI{70}{\degree}$) in 1.8, 4.4, 7.0, 7.6, 9.6, and 11.3 seconds, respectively.



\section{Results and Discussion}

\begin{figure*}
    \centering
    \includegraphics[width=\linewidth]{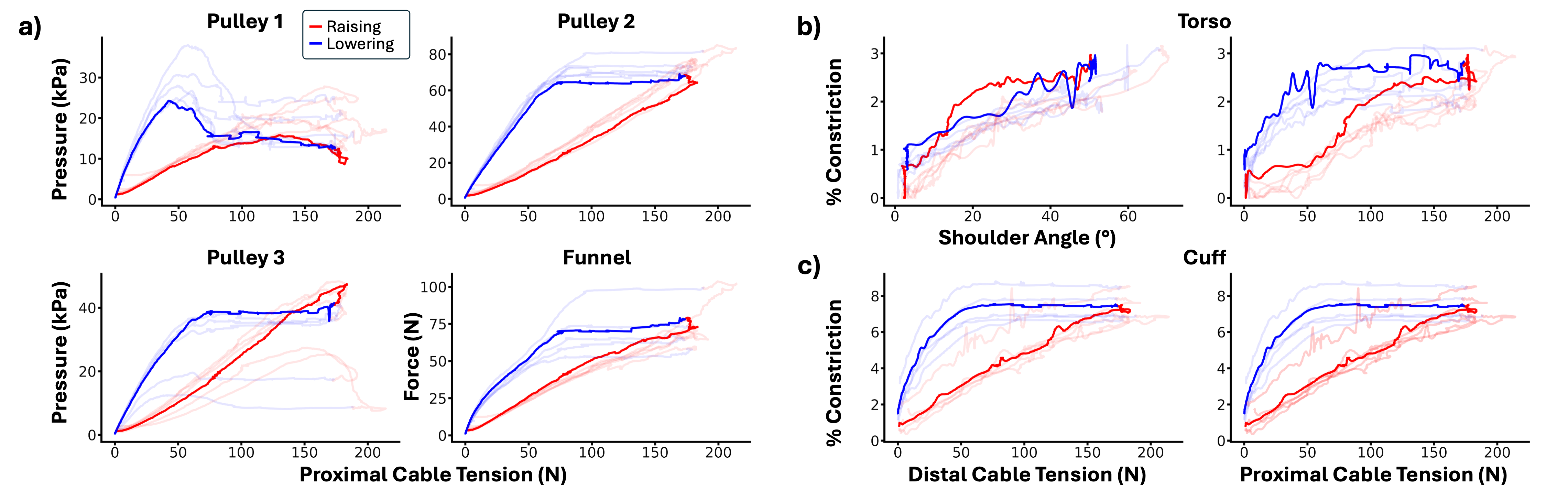}
    \caption{\textit{Elevate} - Body Interaction. \textbf{a)} Plots of pressure on the shoulder at the pulleys and force at the funnel, against input (proximal) cable tension from the motor. \textbf{b)}  Plots of the percentage reduction in area estimated from the four torso markers vs shoulder angle and vs proximal cable tension. \textbf{c)} Plots of the percentage reduction in area estimated from the four cuff markers vs distal cable tension and vs proximal cable tension. 
    }
    \label{fig:body interaction}
\end{figure*}

The results of the experimental evaluation are presented in Fig.\,\ref{fig:body interaction}. As discussed previously, six trials with varying rates of torque change were performed to understand interaction forces under varying dynamics. The median trial is presented in solid colour while the remaining trials are plotted at a higher transparency. The results present estimations of the the pressure experienced at the pulley base and the force at the funnel (Fig.\,\ref{fig:body interaction}a), the constriction estimated at the torso (Fig.\,\ref{fig:body interaction}b), and the constriction estimated at the upper arm (Fig.\,\ref{fig:body interaction}c). These estimations are reported against the cable tension at the motor (proximal), with the constriction at the cuff is also plotted against the cable tension at the cuff (distal). The reported data provides insight into how suit-body interactions vary with changes in motor torque and cable tension, and serves as a benchmark for future improvements in balancing suit ergonomics against effective assistance provided by the \textit{Elevate} suit.

\subsection{Pressure on the Shoulder}

\textit{Elevate} was designed so that, where possible, the loads on the body were normal rather than shear forces. This design consideration is primarily seen in the three pulleys over the shoulder where the exposed cable and low friction bearings remove the shear force seen when using Bowden cables. As expected, the middle pulley exerted the highest pressures on the shoulder due to the greater change in cable direction, with estimated pressures ranging from $\SI{69.0}{\kilo\pascal}$ to $\SI{85.1}{\kilo\pascal}$ across all trials, which is in the range of pressures exerted by our fingers while pressing on an object \cite{gurram_study_1995}. The foam under the shoulder strap meant that this pressure was spread over a larger area and so the true values were lower than reported, with the wearer not reporting discomfort. If required, the suit can be modified to further reduce this pressure, for example, by increasing the number of pulleys to further reduce the change of direction in the cable allowing for lower localised pressure. The maximum force on the body at the re-routing funnel is approximately $\SI{6}{\kilo\gram}$ of pulling force. Due to the suit geometry, this force is not normal to the body and is distributed through padding and the straps over the left shoulder. No discomfort was reported by the wearer. This loading could again be reduced by modifying the placement of the redirect funnel however this would affect the moment arm of the cable, reducing the maximum effective assistance of the suit.

\subsection{Compression across the Torso and Upper Arm }

Fig.\,\ref{fig:body interaction}b\&c present the estimated compression of the torso and upper arm under applied cable tension. Even with  $\approx\SI{200}{\newton}$ of cable tension measured at the motor, the resulting grounding forces produced $\leq3\%$ constriction at the torso. Owing to transmission losses, the corresponding distal cable tension was reduced to  $\approx\SI{120}{\newton}$ but still resulted in cuff constriction of $\leq8\%$. Notably, the tightly grouped and consistent profiles observed across varying assistive torque rates indicate that the suit maintains stable interaction with the wearer despite changes in cable tension and dynamics. This stability is favourable for wearer comfort and ergonomics, and results in more predictable suit behaviour, thus improving controllability.

As expected, soft tissue and suit deformation give rise to pronounced hysteretic behaviour that is seen across all measurements (Fig.\,\ref{fig:body interaction}) and is most notably displayed in the estimated cuff constriction plotted against proximal and distal cable tension (Fig.\,\ref{fig:body interaction}c). These results indicate that substantial losses occur between the motor and cuff due to material and soft tissue deformation. 

\subsection{Limitations and Future Work}
The focus of this work was an evaluation of the non-assistive forces between the body and the \textit{Elevate} suit, as such, we did not attempt to estimate assistance or metabolic cost. Hence, we do not claim that the suit provides $\SI{100}{\percent}$ assistance as some involuntary activation of the shoulder muscles may have occurred in response to the action of the suit. However, our supplementary videos clearly show \textit{Elevate}'s ability for full shoulder assistance. Future work will include muscle electromyography measurements. Additionally, we moved the actuation unit off-board to allow placement of the proximal force sensor and to study force transmission across the suit and body. A limitation of this setup is the significant friction and losses introduced by the sharp bend in the Bowden tube caused by the horizontal motor mounting. Future work will focus on measuring and optimising power transmission efficiency to further improve comfort, ergonomics, and effective assistance.

In total, the participant wore the suit for over eight hours across two sessions and reported no pain or discomfort with only minor marking of the skin at the upper arm. To fully validate \textit{Elevate}'s adaptability and comfort, future work will include trials with a greater range of participants, varying in sex and body type, combined with a qualitative survey. Finally, motor torque was limited to $\approx\SI{4.2}{\newton\meter}$ as the motor used was not backdrivable. In our next study, we aim to rectify this limitation by incorporating an intuitive controller that allows us to safely achieve higher assistive torques.

\section{Conclusion}
This work presents an in-depth experimental evaluation of the suit-body interaction of the \textit{Elevate} soft robotic suit for shoulder elevation. Through the presented methodology, we propose a quantitative evaluation of suit-body interaction that does not require expensive pressure-sensing mats, and instead leverages measurements of geometry, kinematics and cable tension. We demonstrated the suit's ability to transmit $\approx\SI{120}{\newton}$ of cable tension to the cuff while only resulting in pressures equivalent to forces from a hand grasp ($\approx\SI{69.1-85.1}{\kilo\pascal}$) over the shoulder and $\leq3\%$ and $\leq8\%$ of constriction around the torso and upper-arm, respectively. This evaluation validates the design philosophy of \textit{Elevate} in delivering effective assistance without compromising on stability and ergonomics. Furthermore, the study will also act as a benchmark to guide future development of \textit{Elevate}. We hope these results will be used by future researchers to benchmark alternative SRS designs and encourage them to perform similar evaluations that go beyond measuring assistive torques and metabolic cost savings in order to realise more effective and comfortable suits capable of lab-to-real-world translation.

\section*{Acknowledgments}
The authors would like to thank Isabella Szczech for her design inputs and assistance during data collection. This work utilised prototyping equipment at the Imperial College Advanced Hackspace. The authors would like to thank the mentors at the Hackspace for their support.

\bibliographystyle{IEEEtran}
\bibliography{paper1}

@article{xiloyannis_soft_2022,
	title = {Soft {Robotic} {Suits}: {State} of the {Art}, {Core} {Technologies}, and {Open} {Challenges}},
	volume = {38},
	issn = {1941-0468},
	shorttitle = {Soft {Robotic} {Suits}},
	doi = {10.1109/TRO.2021.3084466},
	number = {3},
	journal = {IEEE Transactions on Robotics},
	author = {Xiloyannis, Michele and Alicea, Ryan and Georgarakis, Anna-Maria and Haufe, Florian L. and Wolf, Peter and Masia, Lorenzo and Riener, Robert},
	month = jun,
	year = {2022},
	pages = {1343--1362},
}

@article{tiffreau_pain_2006,
	title = {Pain and {Neuromuscular} {Disease}: {The} {Results} of a {Survey}},
	volume = {85},
	issn = {0894-9115},
	shorttitle = {Pain and {Neuromuscular} {Disease}},
	doi = {10.1097/01.phm.0000228518.26673.23},
	language = {en-US},
	number = {9},
	urldate = {2025-04-04},
	journal = {American Journal of Physical Medicine \& Rehabilitation},
	author = {Tiffreau, Vincent and Viet, Ghislaine and Thévenon, André},
	month = sep,
	year = {2006},
	pages = {756},
}

@article{qu_observation_2025,
	title = {Observation of risk factors for shoulder subluxation after stroke using ultrasonography to measure thickness of the supraspinatus muscle: a cross-sectional study},
	volume = {16},
	issn = {1664-2295},
	shorttitle = {Observation of risk factors for shoulder subluxation after stroke using ultrasonography to measure thickness of the supraspinatus muscle},
	doi = {10.3389/fneur.2025.1532004},
	language = {eng},
	journal = {Frontiers in Neurology},
	author = {Qu, Yu and Shi, Xiue and Wang, Yongyong and Ji, Tiecheng and Chen, Lei and Yu, Suli and Huo, Ming},
	year = {2025},
	pmid = {40177404},
	pmcid = {PMC11961407},
}

@incollection{varghese_wearable_2018,
    title = {Wearable {Robotics} for {Upper}-{Limb} {Rehabilitation} and {Assistance}},
    isbn = {978-0-12-811810-8},
    booktitle = {Wearable {Technology} in {Medicine} and {Health} {Care}},
    author = {Varghese, Rejin and Freer, Daniel and Deligianni, Fani and Liu, Jindong and Yang, Guang-Zhong},
    month = jan,
    year = {2018},
    doi = {10.1016/B978-0-12-811810-8.00003-8},
    pages = {23--69},
}

@article{lee_effectiveness_2022,
    title = {Effectiveness of {Rehabilitation} {Exercise} in {Improving} {Physical} {Function} of {Stroke} {Patients}: {A} {Systematic} {Review}},
    volume = {19},
    copyright = {http://creativecommons.org/licenses/by/3.0/},
    issn = {1660-4601},
    shorttitle = {Effectiveness of {Rehabilitation} {Exercise} in {Improving} {Physical} {Function} of {Stroke} {Patients}},
    doi = {10.3390/ijerph191912739},
    language = {en},
    number = {19},
    urldate = {2025-04-30},
    journal = {International Journal of Environmental Research and Public Health},
    author = {Lee, Kyung Eun and Choi, Muncheong and Jeoung, Bogja},
    month = jan,
    year = {2022},
    publisher = {Multidisciplinary Digital Publishing Institute},
    pages = {12739},
}

@article{bos_prevalence_2019,
    title = {The prevalence and severity of disease-related disabilities and their impact on quality of life in neuromuscular diseases},
    volume = {41},
    issn = {0963-8288},
    doi = {10.1080/09638288.2018.1446188},
    number = {14},
    urldate = {2025-04-30},
    journal = {Disability and Rehabilitation},
    author = {Bos, Isaac and , Klaske, Wynia and , Josué, Almansa and , Gea, Drost and , Berry, Kremer and and Kuks, Jan},
    month = jul,
    year = {2019},
    pmid = {29514523},
    publisher = {Taylor \& Francis},
    pages = {1676--1681},
}

@inproceedings{xiloyannis_characterisation_2018,
    title = {Characterisation of {Pressure} {Distribution} at the {Interface} of a {Soft} {Exosuit}: {Towards} a {More} {Comfortable} {Wear}},
    isbn = {978-3-030-01887-0},
    shorttitle = {Characterisation of {Pressure} {Distribution} at the {Interface} of a {Soft} {Exosuit}},
    doi = {10.1007/978-3-030-01887-0_7},
    language = {en},
    booktitle = {Wearable {Robotics}: {Challenges} and {Trends}},
    author = {Xiloyannis, Michele and Chiaradia, Domenico and Frisoli, Antonio and Masia, Lorenzo},
    editor = {Carrozza, Maria Chiara and Micera, Silvestro and Pons, José L.},
    year = {2018},
    pages = {35--38},
}

@article{piao_development_2023,
    title = {Development of a comfort suit-type soft-wearable robot with flexible artificial muscles for walking assistance},
    volume = {13},
    copyright = {2023 The Author(s)},
    issn = {2045-2322},
    language = {en},
    number = {1},
    urldate = {2026-01-19},
    journal = {Scientific Reports},
    author = {Piao, Jiaoli and Kim, Minseo and Kim, Jeesoo and Kim, Changhwan and Han, Seunghee and Back, Inryeol and Koh, Je-sung and Koo, Sumin},
    month = mar,
    year = {2023},
    pages = {4869},
}

@article{yandell_physical_2017,
    title = {Physical interface dynamics alter how robotic exosuits augment human movement: implications for optimizing wearable assistive devices},
    volume = {14},
    issn = {1743-0003},
    shorttitle = {Physical interface dynamics alter how robotic exosuits augment human movement},
    language = {en},
    number = {1},
    urldate = {2026-01-19},
    journal = {Journal of NeuroEngineering and Rehabilitation},
    author = {Yandell, Matthew B. and Quinlivan, Brendan T. and Popov, Dmitry and Walsh, Conor and Zelik, Karl E.},
    month = may,
    year = {2017},
    pages = {40},
}

@article{gurram_study_1995,
    title = {A study of hand grip pressure distribution and {EMG} of finger flexor muscles under dynamic loads},
    volume = {38},
    issn = {0014-0139},
    number = {4},
    urldate = {2026-01-23},
    journal = {Ergonomics},
    author = {Gurram, R. and Rakheja, S. and Gouw, G. J.},
    month = apr,
    year = {1995},
    pmid = {7729396},
    pages = {684--699},
}

\vfill

\end{document}